\documentclass[journal]{IEEEtran}
\usepackage{amsmath,amsfonts}
\usepackage{algorithmic}
\usepackage{algorithm}
\usepackage{array}
\usepackage{booktabs}
\usepackage[table]{xcolor}
\usepackage{makecell}
\usepackage[caption=false,font=normalsize,labelfont=sf,textfont=sf]{subfig}
\usepackage{textcomp}
\usepackage{stfloats}
\usepackage{tabularx}
\usepackage{url}
\usepackage{verbatim}
\usepackage{graphicx}
\usepackage{multirow}
\usepackage[colorlinks,urlcolor=blue,linkcolor=blue,citecolor=blue]{hyperref}
\usepackage{pifont}

\usepackage{orcidlink}
\newcommand{\cmark}{\ding{51}} 
\newcommand{\xmark}{\ding{55}} 
\DeclareUnicodeCharacter{FF1A}{:}

\usepackage[numbers]{natbib}
\newcolumntype{Y}{>{\raggedright\arraybackslash}X}

\begin{document}

\title{Beyond Static Models: An Evolving Framework for Continual Learning in Large Language Models across Training Stages}

\author{Hongyang Chen\,\orcidlink{0000-0002-7626-0162},~\IEEEmembership{Senior Member,~IEEE,} Zhongwu Sun, Hongfei Ye, Kunzhi Li, Xuemin Lin\,\orcidlink{0000-0003-2396-7225},~\IEEEmembership{Fellow,~IEEE}%
\thanks{H. Chen is with the Zhejiang Lab, Hangzhou 311100, China (email: dr.h.chen@ieee.org; hongyang@zhejianglab.com).}%
}

\markboth{Journal of \LaTeX\ Class Files,~Vol.~xx, No.~x, September ~2025}%
{H. Chen \MakeLowercase{\textit{et al.}}: Continual Learning in Large Language Models across Training Stages}



\maketitle

\begin{abstract}
Continual learning (CL) has emerged as a pivotal paradigm to enable large language models (LLMs) to dynamically adapt to evolving knowledge and sequential tasks while mitigating catastrophic forgetting, a critical limitation of the static pre-training paradigm inherent to modern LLMs. This paper presents a comprehensive overview of CL methodologies tailored for LLMs, structured around three core training stages: continual pre-training, continual fine-tuning, and continual alignment. Beyond the canonical taxonomy of rehearsal-, regularization-, and architecture-based methods, we further subdivide each category by its distinct forgetting mitigation mechanisms and conduct a rigorous comparative analysis of the adaptability and critical improvements of traditional CL methods for LLMs. In doing so, we explicitly highlight core distinctions between LLM CL and traditional machine learning, particularly with respect to scale, parameter efficiency, and emergent capabilities. Our analysis covers essential evaluation metrics, including forgetting rates and knowledge transfer efficiency, along with emerging benchmarks for assessing CL performance. While current methods show promising results, fundamental challenges persist in achieving seamless knowledge integration, and we identify key open problems and future directions for the field.
\end{abstract}


\section{Introduction}
Since the introduction of the Transformer architecture \cite{attention}, a plethora of Transformer-based pre-trained language models have been proposed and widely adopted, exemplified by seminal works such as BERT \cite{bert}, GPT \cite{gpt3}, and LLaMA \cite{llama}. By undergoing massive-scale pre-training on diverse corpora, these large language models (LLMs) exhibit emergent capabilities, which serve as direct empirical evidence for the scaling law—the principle that model performance and generalization improve systematically with increases in parameter count and training data volume. As a result, LLMs have achieved state-of-the-art performance across a wide range of natural language processing (NLP) tasks, demonstrating strong generalization, in-context learning, and instruction-following abilities, thereby establishing a transformative paradigm for modern NLP research \cite{llmsurvey}. 

However, the success of LLMs is inherently contingent upon a static pre-training paradigm, which stands in sharp contrast to the dynamic, ever-evolving nature of real-world scenarios. First, the real world is characterized by perpetual flux: new knowledge, concepts, events, and linguistic usages emerge incessantly, rendering it infeasible to curate a single, exhaustive dataset that encapsulates the full spectrum of potential information. Second, stringent data privacy regulations and ethical constraints preclude the collection of sensitive or proprietary data, further undermining the completeness and comprehensiveness of pre-training corpora. Third, re-pre-training LLMs on updated data is prohibitively expensive— the massive scale of both models and datasets incurs substantial computational and financial burdens, rendering full retraining impractical for frequent knowledge updates. In contrast, the human brain exhibits a remarkable capacity for incremental learning: it acquires new knowledge without catastrophic forgetting of prior skills, while adapting existing knowledge to align with novel situations. Inspired by this biological intelligence, there is an urgent imperative for large language models (LLMs) to emulate such a capability—specifically, to incrementally learn from new data or tasks, retain pre-acquired knowledge, and dynamically adapt to evolving real-world scenarios. This critical misalignment between LLMs’ static pre-training paradigm and the inherent dynamicity of real-world knowledge has firmly established continual learning for LLMs as a pivotal, high-priority research direction.

As a well-established paradigm across machine learning (ML), computer vision (CV), and natural language processing (NLP) \cite{Ke2023}, continual learning enables neural networks to sequentially learn a sequence of domain corpora or end-tasks, and for LLMs, it has emerged as a promising solution to incorporate new knowledge at minimal cost while preserving existing expertise. Empirical evidence \cite{Gururangan2020} corroborates that applying continual learning methodologies to pre-trained LLMs not only updates their internal knowledge repositories but also substantially boosts task-specific performance. Compared to training from scratch, continual learning leverages pre-acquired knowledge to drastically reduce computational overhead and improve training efficiency, while sustaining state-of-the-art performance—this unique synergy of efficiency and adaptability lays a robust foundation for LLM deployment in real-world scenarios, directly addressing the core challenge of static pre-training amid dynamic knowledge evolution.

Several surveys on LLM continual learning (CL) already exist \cite{clsurvey1, clsurvey2, guo2025comprehensive}, but they exhibit critical limitations. Specifically, \cite{clsurvey1} partitions LLM CL into three stages (Continual Pre-Training, Domain-Adaptive Pre-training, Continual Fine-Tuning) and analyzes methods across domains (e.g., medical, legal, financial) alongside core challenges (e.g., language shift, content shift), yet lacks fine-grained categorization of CL methods by their underlying mechanisms. Similarly, \cite{clsurvey2} adopts a three-stage division but focuses on application scenarios rather than methodological types or catastrophic forgetting mitigation mechanisms. While \cite{guo2025comprehensive} delivers a fine-grained methodological analysis of LLM CL, it neither partitions the CL process into distinct training stages nor focuses exclusively on LLMs, instead encompassing a broader spectrum of generative AI models (e.g., multimodal models, diffusion models).

Distinct from these works, this paper focuses exclusively on continual learning for large language models (LLMs). We structure the discussion around three dedicated stages—continual pre-training, continual fine-tuning, and continual alignment—and provide a systematic analysis of representative methodologies within each stage. Beyond the canonical taxonomy of traditional CL (i.e., rehearsal-based, regularization-based, and architecture-based methods), this paper further interprets existing methods through unified lenses of knowledge dynamics, stage-specific constraints, and efficiency-effectiveness trade-offs.

Accordingly, this paper makes three main contributions. First, under the theme of \textit{Beyond Static Models}, it frames the central problem of LLM continual learning as the transition from static pre-training to continual knowledge evolution. Second, as \textit{An Evolving Framework}, it moves beyond method cataloging by introducing an analytical perspective based on collision, isolation, and fusion of knowledge representations. Third, \textit{Across Training Stages}, it places continual pre-training, continual fine-tuning, and continual alignment within a common coordinate system, clarifying both their shared goals and their distinct optimization, resource, and alignment challenges.

\noindent\textbf{Survey scope and methodology.} This review focuses on representative studies on continual learning for large language models published between 2020 and early 2026. The literature was primarily collected from Google Scholar, arXiv, ACL Anthology, and IEEE Xplore using keywords such as ``continual learning,'' ``lifelong learning,'' ``large language model,'' ``continual pre-training,'' ``continual fine-tuning,'' and ``continual alignment.'' We prioritized studies that explicitly address sequential adaptation, forgetting mitigation, knowledge transfer, or preference evolution in LLM settings, and we organize the selected papers according to training stage and underlying mitigation mechanism.


\section{Preliminaries}

\begin{figure*}[t]
\vspace{-10pt}
\begin{center}
  \includegraphics[width=\textwidth]{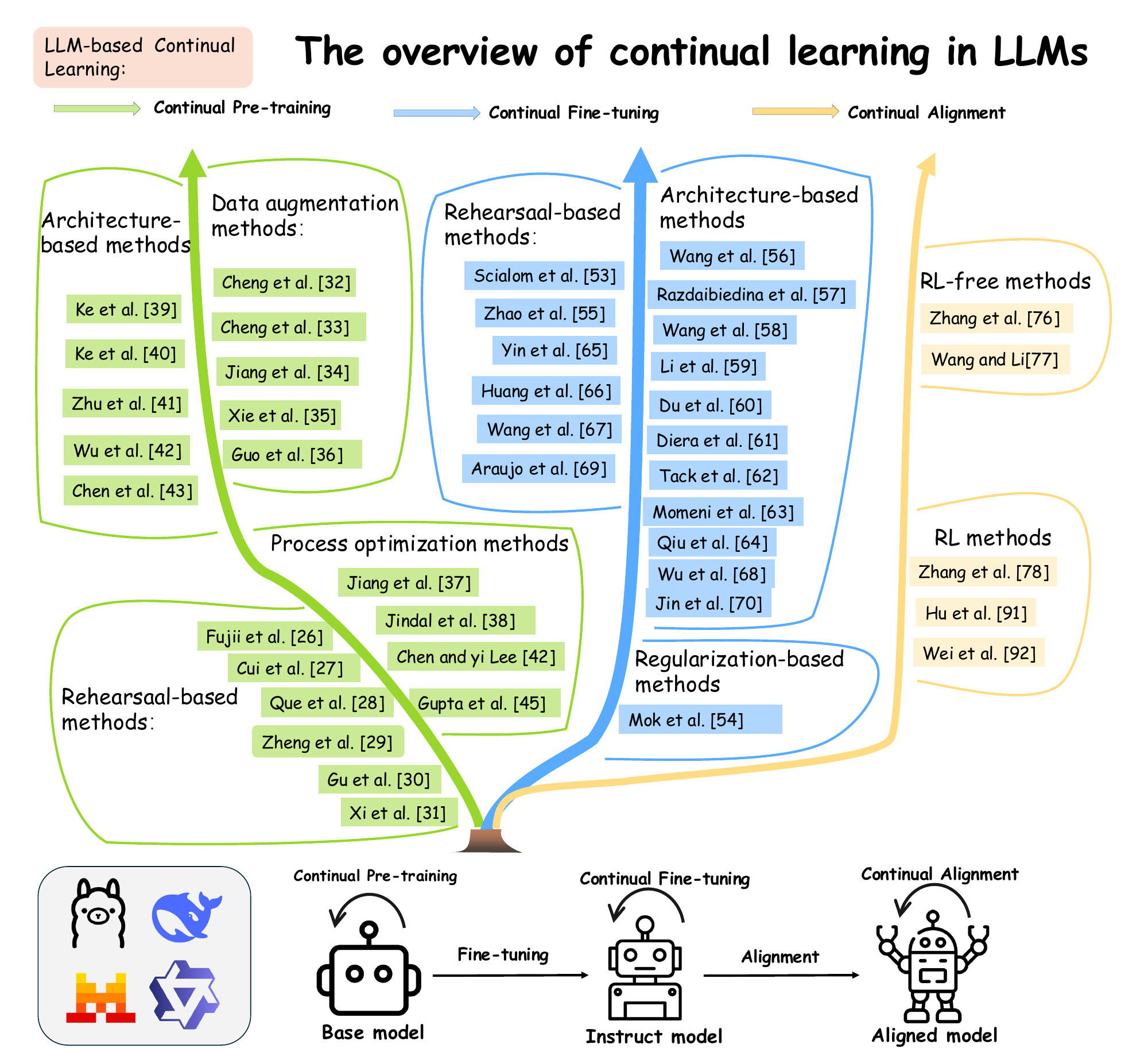} 
  \caption{A schematic overview of continual learning (CL) methodologies for large language models (LLMs). The outermost ring shows the three training stages (continual pre-training, continual fine-tuning, and continual alignment); inner nodes show method families within each stage (e.g., rehearsal, regularization, architecture/PEFT, RL-free, and RL-based); cited works appear as leaf nodes. Color coding indicates the primary knowledge-dynamic mode: collision mitigation (red family), isolation (green family), or fusion (purple family). Arrows indicate the sequential dependency between stages.}
  \label{fig:1} 
\end{center}
\end{figure*}

\subsection{Large language models}
Following the success of foundational models like GPT\cite{gpt3} and LLaMA\cite{llama}, even more powerful large language models (LLMs) have emerged, such as Claude\cite{claude3}, Qwen\cite{qwen25}, and DeepSeek\cite{deepseekv3}, further accelerating the advancement of natural language processing. The rapid development of LLMs has drawn significant attention, largely due to their massive parameter counts and impressive emergent capabilities. The training of LLMs primarily involves three stages: pre-training, fine-tuning, and alignment. \par
In the \textit{pre-training} stage, large language models (LLMs) are trained in a self-supervised manner on extensive corpora, which typically encompass diverse text sources such as books, websites, academic papers, and other publicly accessible content \cite{hu2023survey}. The primary objective of this stage is to equip the model with a foundational understanding of linguistic structures and semantics by learning to predict subsequent tokens in a sequence. Specifically, the optimization goal during pre-training often revolves around minimizing a loss function associated with next-token prediction, such as the cross-entropy loss, as defined in~\eqref{eq:pretrain-loss}:  
\begin{equation}
\label{eq:pretrain-loss}
\mathcal{L} = -\frac{1}{T} \sum_{t=1}^{T} \log P(w_t | w_1, w_2, ..., w_{t-1})
\end{equation}
where \( T \) represents the length of the sequence, \( w_t \) is the token at time step \( t \), and \( P(w_t | w_1, w_2, ..., w_{t-1}) \) denotes the predicted probability of the token \( w_t \) given the preceding context. Through this pre-training process, LLMs acquire an understanding of grammatical rules and accumulate general knowledge. However, as base models, they are primarily limited to text sequence completion and lack capabilities for instruction-following or engaging in conversational interactions \cite{dong2019unified}. \par
In the \textit{fine-tuning} stage, large language models (LLMs) are further refined using high-quality instruction datasets \cite{zhang2023instruction}. This stage not only enhances the model's performance on specific tasks but also improves its ability to follow instructions, making it more reliable and versatile for practical applications. Among the various fine-tuning approaches, Supervised Fine-Tuning (SFT) is the most widely adopted method. Many open-source LLMs undergo SFT on small, domain-specific datasets to adapt them for downstream applications in specialized fields.  
The optimization goal during supervised fine-tuning typically focuses on aligning the model's outputs with human-provided instructions or desired responses. This is achieved by minimizing the cross-entropy loss (see~\eqref{eq:sft-loss}), which measures the divergence between the model's predictions and the ground-truth responses in the instruction dataset:
\begin{equation}
\label{eq:sft-loss}
\mathcal{L}_{\text{SFT}} = -\frac{1}{N} \sum_{i=1}^{N} \log P(y_i | x_i, \theta),
\end{equation}
where \( N \) denotes the number of instruction-response pairs in the dataset, \( x_i \) represents the input instruction or context for the \( i \)-th sample, \( y_i \) is the corresponding target response, and \( P(y_i | x_i, \theta) \) is the probability of generating the target response given the input and the model parameters \( \theta \). Through this process, the model learns to generate contextually relevant and instruction-aligned outputs, thereby improving its applicability to real-world tasks. \par

Finally, in the \textit{alignment} stage, the models are optimized using human feedback to ensure their outputs align with human preferences and values \cite{wang2023aligning}. This stage, often referred to as Reinforcement Learning with Human Feedback (RLHF), aims to imbue models with human-aligned value systems and preferences. Representative methods include Proximal Policy Optimization (PPO)\cite{ppo} and Direct Preference Optimization (DPO)\cite{dpo}. PPO first trains a reward model \(r_\phi(x, y)\) on human preference data and then fine-tunes the language model \(\pi_\theta(y|x)\) using reinforcement learning to maximize the expected reward, while constraining updates to ensure stability. In contrast, DPO directly optimizes the language model on preference data without an explicit reward model, simplifying the alignment pipeline.
The optimization objectives for these methods are defined as follows:

\paragraph{PPO Objective} PPO optimizes the clipped surrogate objective:
\begin{equation}
\label{eq:ppo}
\begin{aligned}
\mathcal{L}_{\text{PPO}}(\theta)
&= \mathbb{E}_{(x, y) \sim \pi_{\theta_{\text{old}}}} \Big[\\
&\quad \min\Big(r_\theta(y|x) \hat{A}(x, y),\\
&\qquad \mathrm{clip}\big(r_\theta(y|x), 1-\epsilon, 1+\epsilon\big)\hat{A}(x, y)\Big)
\Big]
\end{aligned}
\end{equation}
where \(r_\theta(y|x) = \frac{\pi_\theta(y|x)}{\pi_{\theta_{\text{old}}}(y|x)}\) is the probability ratio, \(\hat{A}(x, y)\) is the estimated advantage function based on the reward model, and \(\epsilon\) is the clipping parameter that limits policy updates.

\paragraph{DPO Objective} DPO optimizes a preference-based likelihood loss derived from pairwise human feedback:
\begin{equation}
\label{eq:dpo}
\begin{aligned}
\mathcal{L}_{\text{DPO}}(\theta)
&= -\mathbb{E}_{(x, y^+, y^-)}\Big[ \\
&\quad \log\sigma\Big(\beta\big(\log \pi_\theta(y^+|x) \\
&\qquad - \log \pi_\theta(y^-|x)\big)\Big)
\Big]
\end{aligned}
\end{equation}
where \((x, y^+, y^-)\) are triplets of prompt, preferred response, and less-preferred response, \(\sigma\) is the sigmoid function, and \(\beta\) is an inverse temperature parameter controlling the sharpness of preference.

This alignment process is critical for mitigating harmful, biased, or unsafe behaviors in language models and for ensuring that their responses are consistent, contextually appropriate, and aligned with human ethical standards.

\subsection{Traditional continual learning}
Continual learning, also known as lifelong learning and incremental learning, is an advanced machine learning algorithm that learns continuously, accumulates knowledge from the past, and adapts it to facilitate future learning \cite{chen2018lifelong}. It aims to train sequentially on a series of tasks, ensuring that performance on previous tasks does not degrade while training on new tasks. The main objectives of continual learning are to avoid catastrophic forgetting and facilitate knowledge transfer. Catastrophic forgetting \cite{French1999} refers to the sharp decline in performance on previous tasks when the model is trained on new tasks. Knowledge transfer includes both forward and backward transfer. Forward transfer means using the knowledge learned from previous tasks to aid in learning new tasks, while backward transfer refers to the improvement of performance on previous tasks through the knowledge obtained from new tasks.

The primary definition of continual learning (CL) is as follows \cite{chen2018lifelong}:  
CL considers a sequence of tasks \(\mathcal{T} = \{1, 2, \ldots, T\}\), where \(T\) denotes the total number of tasks. Each task \(t \in \mathcal{T}\) is associated with a training set \(\mathcal{D}_t = \{(x_i^{(t)}, y_i^{(t)})\}_{i=1}^{N_t}\), where \(x_i^{(t)} \in \mathcal{X}\) is an input sample from the shared input space \(\mathcal{X}\), \(y_i^{(t)} \in \mathcal{Y}_t\) is its corresponding label from the task-specific label space \(\mathcal{Y}_t\), and \(N_t\) denotes the number of samples in task \(t\). The input data of each task follows a task-specific distribution \(p(X_t)\), where \(p(X_t) \ne p(X_{t'})\) for any \(t \ne t'\), reflecting the distributional shift across tasks.  

In the CL setting, the goal is to learn a parameterized model \(f_\theta : \mathcal{X} \rightarrow \mathcal{Y}\) that sequentially adapts to each task \(t\), while retaining satisfactory performance on all previously learned tasks. Formally, this can be expressed as the minimization of the cumulative loss across all tasks:
\begin{equation}
\min_\theta \; \sum_{t=1}^T \mathbb{E}_{(x, y) \sim \mathcal{D}_t} \left[ \mathcal{L}\big(f_\theta(x), y\big) \right],
\label{eq:cl-objective}
\end{equation}
subject to constraints that mitigate catastrophic forgetting and promote knowledge transfer. Here, \(\mathcal{L}\) denotes the task-specific loss function (e.g., cross-entropy for classification).

In traditional machine learning, three primary families of approaches have been proposed to instantiate the cumulative objective in~\eqref{eq:cl-objective}, each implementing a different bias on how new tasks should interact with previously consolidated knowledge: rehearsal-based, regularization-based, and architecture-based methods.\par
\textit{\textbf{Rehearsal-based methods}} (a \textit{fusion}-oriented strategy in our terminology) store samples from previous tasks or generate pseudo-samples via a generator to revisit historical support regions during new-task training. They directly approximate the joint loss of~\eqref{eq:cl-objective}, but at the cost of storage burden and unresolved data-privacy concerns when raw historical samples must be retained.

\textit{\textbf{Regularization-based methods}} (a \textit{collision}-mitigation strategy) augment the task loss in~\eqref{eq:cl-objective} with a penalty term that restricts updates to parameters deemed important for prior tasks. HAT \cite{serra2018} introduces task-specific hard attention masks that gate gradient flow between neurons; SPG \cite{konishi2023parameter} relaxes this to soft masks by computing per-unit importance scores, constraining the backward pass while still encouraging knowledge sharing.

\textit{\textbf{Architecture-based methods}} (an \textit{isolation}-oriented strategy) facilitate continual learning by expanding the model architecture, typically introducing task-specific parameters while freezing the original weights. CTR \cite{ke2021achieving} is built upon a pre-trained BERT model and incorporates CL-plugins, where each plugin is a capsule \cite{hinton2011transforming} network equipped with a routing algorithm that identifies and transfers knowledge across tasks. Supsup \cite{wortsman2020supermasks} instead identifies a per-task supermask within a fixed network to generate task-specific subnetworks. Both designs trade additional capacity for stronger isolation, which essentially decouples the per-task losses in~\eqref{eq:cl-objective} at the cost of weakened cross-task transfer.

\begin{figure}[!t]
\centering
\includegraphics[width=\columnwidth]{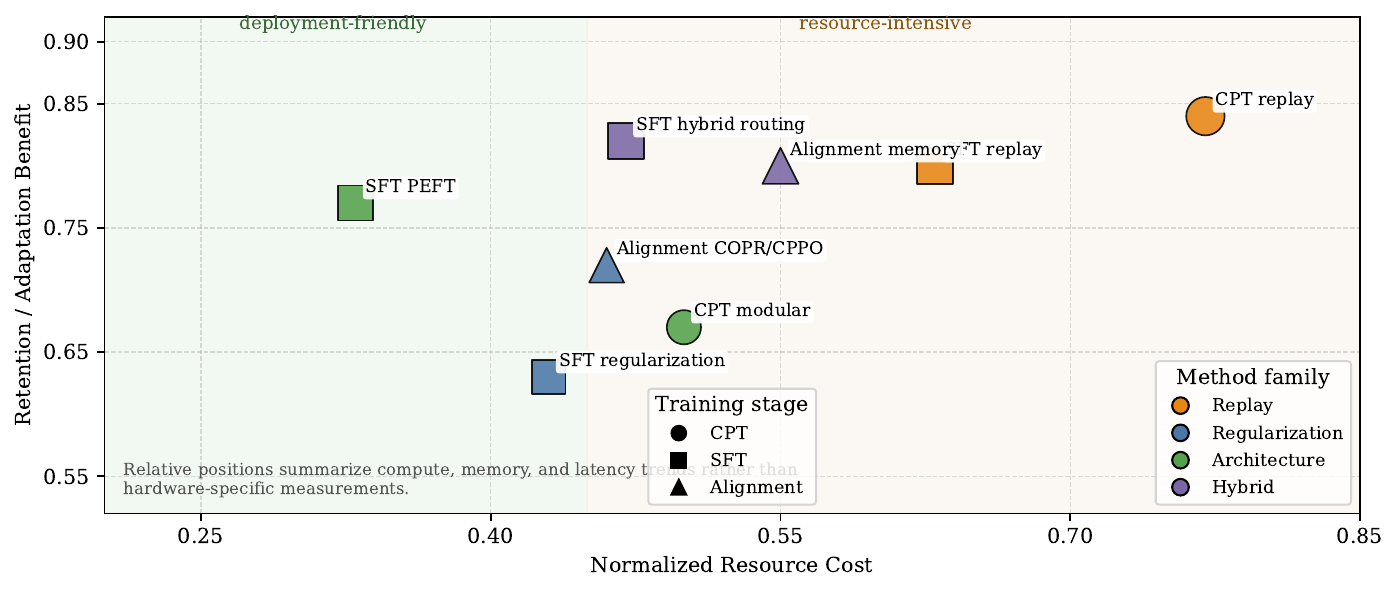}
\caption{Qualitative resource-cost vs.\ retention-benefit map for representative CL methods across the three training stages (CPT, SFT, Align). Methods in the upper-left quadrant are deployment-friendly; axes are normalized for conceptual comparison. Architecture-based and hybrid methods dominate the efficient frontier at the fine-tuning scale, while rehearsal methods remain competitive at the pre-training scale.}
\label{fig:stage-efficiency}
\end{figure}

Moreover, recent LLM-oriented continual learning methods increasingly adopt hybrid modular designs, such as MoE- and PEFT-based approaches, which combine architectural isolation with efficient adaptation. Table~\ref{tab:stage-tradeoff} summarizes the main stage-aware trade-offs among representative continual learning paradigms, highlighting why the same method family can play different roles in continual pre-training, fine-tuning, and alignment. Figure~\ref{fig:stage-efficiency} further visualizes representative methods from all three stages in a two-dimensional resource-cost vs. retention-benefit space, showing that architecture-based and hybrid methods occupy a favorable efficiency frontier at the fine-tuning scale, while rehearsal-based methods remain competitive at the pre-training scale despite higher storage costs.

\begin{table*}[!t]
\caption{Stage-aware qualitative trade-offs among major continual learning paradigms.}
\label{tab:stage-tradeoff}
\centering
\small
\setlength{\tabcolsep}{4pt}
\renewcommand{\arraystretch}{1.0}
\rowcolors{2}{gray!6}{white}
\begin{tabularx}{0.98\textwidth}{@{}p{0.15\textwidth}p{0.18\textwidth}p{0.18\textwidth}p{0.17\textwidth}X@{}}
\toprule
Paradigm & Storage & Training & Inference & Main Advantage \\
\midrule
Rehearsal & High (buffers or stored samples) & Medium (joint replay updates) & No extra latency & Most direct mitigation of forgetting; strong fusion potential \cite{scialom2022fine,fujii2024continual} \\
Regularization & Low (masks or importance statistics) & Medium (constraint-based updates) & No extra latency & Stable parameter budget; works without stored data \cite{kirkpatrick2017overcomingcatastrophicforgettingneural,konishi2023parameter} \\
Architecture-based & Medium (growing task-specific modules) & Low (module-focused updates) & Small extra latency & Strong isolation; near-zero forgetting potential; best suited for fine-tuning scale \cite{wang2023orthogonal,Ke2022} \\
Hybrid (MoE/PEFT) & Medium (experts, adapters, or routing) & Low (sparse PEFT-style updates) & Adaptive (router-dependent) & Scalable LLM adaptation; high flexibility via sparse activation \cite{li2024mixloraenhancinglargelanguage,Zhu2024} \\
\bottomrule
\end{tabularx}
\end{table*}

\subsection{Continual learning in Large Language Models}
LLM continual learning concerns how a model absorbs new knowledge, behaviors, and preferences while preserving previously consolidated capabilities. In contrast to traditional continual learning, the problem is strongly stage-dependent: continual pre-training emphasizes knowledge acquisition under distribution shift, continual fine-tuning focuses on downstream adaptation with minimal capability regression, and continual alignment addresses drifting preferences, safety constraints, and normative consistency. Fig. \ref{fig:1} provides an overall taxonomy, while the following sections examine how these stages differ in mechanism and objective.

\subsection{A Knowledge Dynamics Perspective}
Across training stages, continual learning can be interpreted as regulating knowledge dynamics in a high-dimensional representation space. New updates do not merely append isolated facts; they perturb attention patterns, routing decisions, latent features, and preference policies that already encode historical behavior. The central question is therefore not only how to learn new tasks, but also how new information interacts with previously consolidated knowledge.

We characterize this interaction using three recurring modes. \textit{Collision} occurs when new updates overwrite support regions or decision boundaries that remain important for old knowledge, thereby causing forgetting. \textit{Isolation} allocates new modules, prompts, experts, or memory slots to shield prior representations, improving stability but sometimes weakening transfer. \textit{Fusion} selectively composes old and new knowledge through replay, shared attention, routing, or preference integration, which can improve transfer but also requires stronger control over interference. This view unifies the major method families discussed in this paper: replay revisits old support regions to stabilize overlap; regularization restricts representational drift; modular and PEFT-based methods emphasize structured isolation; and continual alignment methods must additionally reweight competing preference memories instead of treating all historical knowledge as equally reusable.

\begin{figure*}[!t]
\centering
\includegraphics[width=0.94\textwidth]{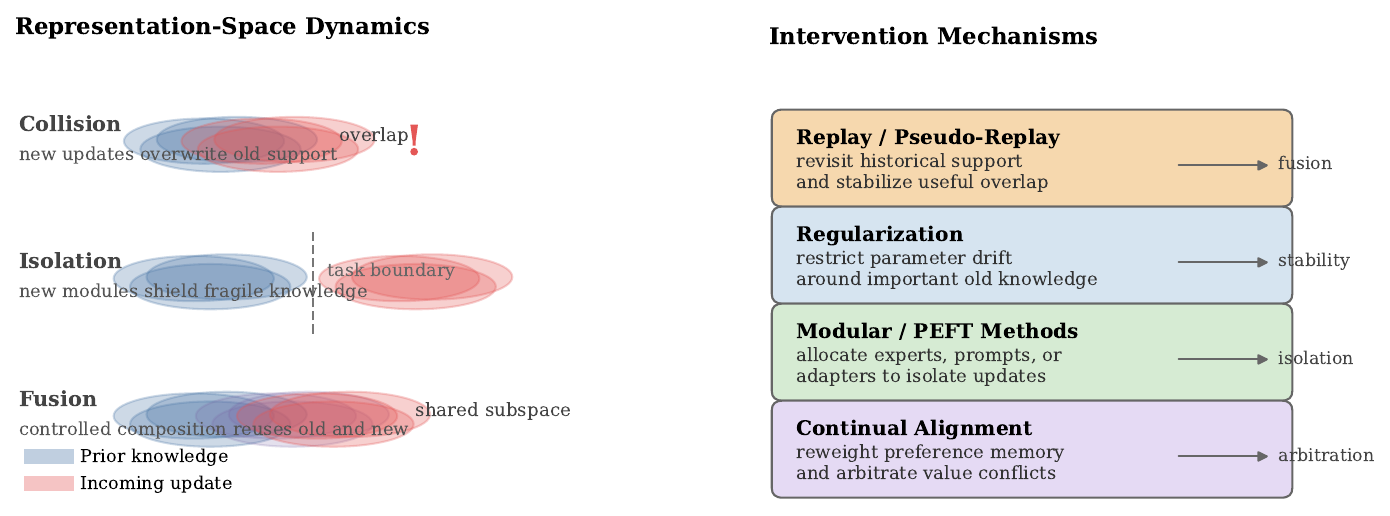}
\caption{Knowledge dynamics and intervention mechanisms in continual learning for LLMs. Left: new updates may collide with, isolate from, or fuse with prior representations. Right: representative method families regulate these dynamics by revisiting historical support, restricting drift, allocating modules, or reweighting preference memories.}
\label{fig:knowledge-dynamics}
\end{figure*}

\section{Continual pre-training}

Continual pre-training updates a pre-trained LLM with new unlabeled corpora, often under domain shift, while attempting to preserve broad linguistic competence. Compared with restarting pre-training from scratch, it offers a more practical route for refreshing factual and domain-specific knowledge under constrained compute \cite{Gururangan2020}. The main risk is catastrophic forgetting of previously learned general-domain capabilities \cite{li2024examiningforgettingcontinualpretraining}.

To analyze this trade-off, this paper organizes continual pre-training methods by the mechanism used to regulate knowledge retention and incorporation: rehearsal, data augmentation, process optimization, and architecture-based adaptation. Table \ref{table1} summarizes representative papers.


\begin{table*}[!t]
\centering
\caption{Representative continual pre-training methods and their main design choices}
\label{table1}
\footnotesize
\renewcommand{\arraystretch}{1.0}
\begin{tabular}{c|cccc|c}
\hline
Paper & Rehearsal & Data augmentation & Process optimization & Architecture-based & Base model \\ \hline
\citet{fujii2024continual} & \cmark & \xmark & \xmark & \xmark & LLaMA2 \\ \hline
\citet{Cui2024} & \cmark & \xmark & \xmark & \xmark & LLaMA-2 7B/13B \\ \hline
\citet{Que2024} & \cmark & \xmark & \xmark & \xmark & Qwen-1.5 \\ \hline
\citet{Zheng2024} & \cmark & \xmark & \xmark & \xmark & LLaMA2 \\ \hline
\citet{Gu2024} & \cmark & \xmark & \xmark & \xmark & LLaMA architecture \\ \hline
\citet{Xi2024} & \cmark & \xmark & \xmark & \xmark & LLaMA-3 8B/70B \\ \hline
\citet{Cheng2024} & \xmark & \cmark & \xmark & \xmark & LLaMA-7B \\ \hline
\citet{Cheng2024a} & \xmark & \cmark & \xmark & \xmark & LLaMA-3 8B \\ \hline
\citet{Jiang2024} & \xmark & \cmark & \xmark & \xmark & LLaMA-3 8B \\ \hline
\citet{Xie2024} & \xmark & \cmark & \xmark & \xmark & Pythia \\ \hline
\citet{Guo2024} & \cmark & \cmark & \xmark & \xmark & OpenLLaMA-3B \\ \hline
\citet{Jiang2024a} & \xmark & \xmark & \cmark & \xmark & LLaMA-2 7B/70B \\ \hline
\citet{Jindal2024a} & \xmark & \xmark & \cmark & \xmark & LLaMA-3/3.1 8B; Qwen2.5 1.5B \\ \hline
\citet{Ke2022} & \xmark & \xmark & \xmark & \cmark & RoBERTa \\ \hline
\citet{Ke2023a} & \xmark & \xmark & \xmark & \cmark & RoBERTa \\ \hline
\citet{Zhu2024} & \xmark & \xmark & \xmark & \cmark & LLaMA-2 7B \\ \hline
\citet{wu2024llamaproprogressivellama} & \xmark & \xmark & \xmark & \cmark & LLaMA-2 7B \\ \hline
\citet{chen2023lifelong} & \xmark & \xmark & \xmark & \cmark & GLaM \\ \hline
\citet{chen2024instructioncpfastapproachtransfer}& \xmark & \xmark & \cmark & \xmark & LLaMA-3 8B \\ \hline
\citet{gupta2023continualpretraininglargelanguage}& \xmark & \xmark & \cmark & \xmark & Pythia \\ \hline
\citet{arannil2024dopaminedomainspecificpretrainingadaptation} & \xmark &  \cmark & \xmark & \xmark & LLaMA-3 8B \\ \hline
\end{tabular}
\end{table*}

\subsection{Rehearsal-based methods}
Rehearsal in CPT operates as a \textit{fusion} mechanism in our framework: by interleaving a subset of prior pre-training data with the new-domain corpus, the optimizer is forced to keep historical support regions covered, so that updates accommodating new tokens cannot freely overwrite the geometry of previously learned representations. The mixed subset acts as a low-cost approximation of the joint loss in~\eqref{eq:cl-objective}. 

Although rehearsal-based methods are conceptually simple, they are widely used in continual pre-training and often remain effective for mitigating catastrophic forgetting, enabling models to retain previously learned information while continually adapting to new domains. 

\citet{fujii2024continual, Cui2024} transfer the base model to other language domains with higher performance than models trained from scratch by utilizing a mixture of new-language data and pre-training data. \citet{fujii2024continual} developed a Japanese language model by continual pre-training LLaMA2 on a mixture of Japanese and English datasets, and demonstrated that model performance improves in line with increases in training data. \citet{Cui2024} continually pre-train LLaMA2 on Chinese data, leading to significant improvements in Chinese text understanding and generation.


Building on prior scaling laws \cite{Kaplan2020, Hoffmann2022}, recent studies have extended these principles to continual pre-training. \cite{Zheng2024} focused on cross-lingual CPT, conducting experiments on 40 model sizes to analyze scaling behaviors across different languages. Meanwhile, \cite{Que2024} investigated domain-specific CPT in specialized fields such as code, math, law, and others, proposing a parameterized domain-continual pre-training (D-CPT) law. \cite{Gu2024} proposed critical mixture ratios for continual pre-training and considered the trade-off between general loss and domain loss.
These works examined the impact of data mixture ratios on downstream task performance across various model sizes and training token scales, offering insights into optimizing data mixing ratios to improve training efficiency and performance under constraints of limited computational and data resources. Collectively, these studies demonstrate that CPT converges faster and requires fewer computational resources compared to training models from scratch.

Similarly, \citet{Xi2024} investigate the optimization of the additional-language mixture ratio and learning rate during the continual pre-training of the large-scale language model LLaMA3-70B in the Chinese domain. The results show that model capability improves not only in the additional language, but also in the original language and even in other specific domains.

The integration of these two data sources enables the model to leverage foundational knowledge acquired during pre-training while adapting to the specific characteristics of the new domain. This approach facilitates a balanced trade-off between retaining prior knowledge and assimilating new information, which is essential for maintaining robust performance in dynamic environments. 

\subsection{Data augmentation methods}
Data augmentation can be read as an indirect \textit{collision-mitigation} strategy: rather than constraining parameters or adding modules, it reshapes the input distribution so that updates triggered by the new corpus are less likely to overwrite the directions in representation space that the original model relies on. High-quality, instruction-rich, or distribution-matched augmentation reduces the magnitude of corrective updates needed during CPT and thereby curbs forgetting at its source.

\citet{Cheng2024} proposed a rule-based data augmentation method to transform raw corpora into reading comprehension texts. Each raw text is enriched with a series of question-answer pairs related to its content. This rule-based method mines intrinsic tasks through a few regex-based patterns and reinforces the model’s ability to respond to prompts effectively while also integrating domain knowledge.
Similarly, \citet{Cheng2024a} trains an instruction synthesizer to convert raw text into question-answer pair instruction data. The synthesized instructions are subsequently combined with the raw text data for continual pre-training.
The Mix-CPT \cite{Jiang2024} decouples knowledge learning and format alignment by combining the raw domain corpus, general instructions, and alignment data to enhance knowledge learning ability. Specifically, it incorporates a self-distillation constraint to alleviate catastrophic forgetting.

\citet{Xie2024} introduces FinPythia-6.9B, which is obtained via continual pre-training in the financial domain. It proposes efficient data selection methods ETS-DACP and ETA-DACP for continual pre-training, which outperform vanilla continual pre-training with just 10\% of the data cost, without any degradation on open-domain standard tasks.

\citet{Guo2024} identify a "stability gap" phenomenon in which initial performance drops before recovering during continual pre-training, and propose three data augmentation strategies to mitigate it. 
They introduce Llama3-Physician, a medical-domain model reported to perform competitively on several medical benchmarks while using only 40\% of the training resources and maintaining general-task performance.

\cite{arannil2024dopaminedomainspecificpretrainingadaptation} introduces an automated framework that uses large language models to generate domain-specific seed data from corpora like Common Crawl. This seed data guides the mining of relevant documents, improving pre-training quality for domain-specific tasks. Techniques include dataset curation, noise filtering, and advanced augmentation to enhance data diversity and relevance.

By enhancing data quality, these data augmentation methods ensure that the model is trained on the most useful and accurate information, enabling models to better generalize and adapt to new tasks and environments.
\subsection{Process optimization methods}
Process-optimization methods sit between collision mitigation and fusion: by reordering or interleaving instruction-tuning with continual pre-training, they restructure when knowledge of different types is consolidated, so that later updates land on a parameter manifold that is already partly aligned with the new objective. This pre-conditioning reduces the destructive interference that vanilla CPT incurs.

PIT \cite{Jiang2024a} argues that instruction data is more straightforward than raw document data, as documents are often more complex and contain factual content in a more intricate manner. To address this, the study proposes pre-instruction-tuning (PIT), which involves QA instruction fine-tuning prior to continual pre-training, to enhance the model's ability to extract factual knowledge across various domains.

\citet{Jindal2024a} examines how continual pre-training impacts instruction-following ability in LLMs, revealing that it can cause catastrophic forgetting when directly continually pre-training instruction-tuned models. It also introduces an efficient strategy to gain up-to-date knowledge and instruction-following capabilities in continual learning.

\cite{chen2024instructioncpfastapproachtransfer} integrates instruction tags into the continual pre-training process, preventing loss of conversational proficiency while acquiring new languages. It requires only 0.1 billion tokens of high-quality instruction-following data, reducing resource consumption.

\citet{gupta2023continualpretraininglargelanguage} investigates learning rate re-warming and re-decaying schedules for continual pre-training, showing that how the optimizer is restarted after a domain switch has a substantial effect on both plasticity and forgetting. Their findings suggest that a short warmup from the previous checkpoint's learning rate is preferable to either a cold restart or continuing the decayed schedule, offering a practical process-level intervention that requires no architectural changes.

\subsection{Architecture-based methods}
Architecture-based CPT methods follow an \textit{isolation} strategy: they expand the model with new modules or route activations via task-specific masks, so that the base weights learned in prior stages are shielded from new-domain gradient flow. Unlike the architectural isolation in fine-tuning (Section~\ref{subsec:cft-arch}), isolation at the pre-training scale must operate over self-supervised objectives without explicit task boundaries, which makes routing and masking substantially harder to calibrate.

\citet{Ke2022} proposes a parallel adapter method that integrates CL-plugins in each transformer block, utilizing a special task mask mechanism to mitigate catastrophic forgetting.
\citet{Ke2023a} introduces DGA, which uses a proxy KL-divergence loss with different dropout percentages to calculate the importance of different attention heads in the transformer block. It also employs a soft-masking mechanism to preserve important knowledge while adapting to a new domain. Similarly, \citet{Ke2023b} explores DAS, extending DGA to all neural units, and investigates the adaptation of a sequence of new domain knowledge by incorporating contrastive learning.

The Mixture-of-Experts (MoE) architecture \cite{shazeer2017outrageouslylargeneuralnetworks, lepikhin2020gshardscalinggiantmodels} has gained increasing popularity for scaling model parameters due to its sparse activation properties. However, training an MoE model from scratch is often constrained by data hunger and training instability. \citet{Zhu2024} proposes LLaMA-MoE, transforming LLaMA2 into an MoE architecture and investigating various expert construction methods and data sampling strategies during continual pre-training. \citet{chen2023lifelong} takes a complementary MoE approach by training distribution-specialized experts, where each expert is dedicated to a specific data distribution encountered during lifelong pre-training; routing incoming tokens to the appropriate specialist preserves prior knowledge without requiring a fixed replay buffer.

\citet{wu2024llamaproprogressivellama} proposes LLaMA Pro, which expands a pre-trained LLaMA by inserting new transformer blocks while freezing the original layers. Only the newly added blocks are trained on the new domain corpus, providing a structural isolation that prevents gradient flow from touching the base model's learned representations. This block-expansion strategy achieves strong domain adaptation with minimal forgetting, and the expanded architecture can be merged back into a single model for deployment.

\subsection{Discussion: take-aways for continual pre-training}
\label{subsec:cpt-discussion}
Synthesizing the four method families above, we offer three observations. \textit{First}, data-mixture rehearsal has become a strong baseline: across \cite{fujii2024continual, Cui2024, Que2024, Zheng2024, Gu2024, Xi2024}, mixing 5–15\% of prior-domain data with new corpora consistently preserves general capabilities, though the optimal ratio remains task- and model-dependent. \textit{Second}, architecture-based methods remain underused at the pre-training scale, due to the cost of inserting and warming up new modules over trillion-token corpora; MoE-style conversions \cite{Zhu2024} are among the few architectural routes with credible CPT-scale evidence. \textit{Third}, the "stability gap" \cite{Guo2024}—an initial performance dip before recovery—remains theoretically underexplained, suggesting that CPT update geometry may differ qualitatively from fine-tuning.

\section{Continual fine-tuning}

\begin{table*}[t]
\centering
\caption{Representative continual fine-tuning methods and their main design choices}
\label{table2}
\begin{tabular}{c|ccc|c}
\hline
\multicolumn{1}{c|}{Paper} & Rehearsal & Regularization-based & Architecture-based & \multicolumn{1}{c}{Base model} \\ \hline
\multicolumn{1}{c|}{\citet{scialom2022fine}} & \cmark & \xmark & \xmark & \multicolumn{1}{c}{T0} \\ \hline
\multicolumn{1}{c|}{\citet{mok2023large}} & \cmark & \cmark & \xmark & \multicolumn{1}{c}{BART} \\ \hline
\multicolumn{1}{c|}{\citet{zhao2024sapt}} & \cmark & \xmark & \cmark & \multicolumn{1}{c}{T5, LLaMA-2} \\ \hline
\multicolumn{1}{c|}{\citet{wang2023orthogonal}} & \xmark & \xmark & \cmark & \multicolumn{1}{c}{T5-large} \\ \hline
\multicolumn{1}{c|}{\citet{razdaibiedina2023}} & \xmark & \xmark & \cmark & \multicolumn{1}{c}{T5, BERT} \\ \hline
\multicolumn{1}{c|}{\citet{wang2023rehearsal}} & \xmark & \xmark & \cmark & \multicolumn{1}{c}{BERT} \\ \hline
\citet{li2024mixloraenhancinglargelanguage} & \xmark & \xmark & \cmark & LLaMA-2 \\ \hline
\citet{du2024unlockingcontinuallearningabilities} & \xmark & \xmark & \cmark & LLaMA-2 \\ \hline
\citet{diera2024continuallearningencoderonlylanguage} & \xmark & \xmark & \cmark & BERT \\ \hline
\citet{tack2024onlineadaptationlanguagemodels} & \xmark & \xmark & \cmark & LLaMA-2 \\ \hline
\citet{momeni2024continuallearningusingkernelbased} & \xmark & \xmark & \cmark & BART \\ \hline
\citet{qiu2024continuallearningusinglarge} & \xmark & \xmark & \cmark & Mistral-7B, LLaMA-3.1-8B \\ \hline
\citet{yin2022contintin} & \cmark & \xmark & \xmark & BART \\ \hline
\citet{huang2024mitigating} & \cmark & \xmark & \xmark & LLaMA-2-7B \\ \hline
\citet{wang2024inscl} & \cmark & \xmark & \xmark & LLaMA-7B \\ \hline
\citet{wu2024switchcitswitchingcontinualinstruction} & \xmark & \xmark & \cmark & BLOOMZ-7.1B \\ \hline
\citet{araujo2024learningroutedynamicadapter} & \cmark & \xmark & \cmark & BERT \\ \hline
\citet{JIN2025112750}& \xmark & \xmark & \cmark & Flan-T5 \\ \hline
\end{tabular}
\end{table*}

Continual fine-tuning incrementally adapts an instruction-tuned model to a sequence of downstream tasks while retaining previously acquired capabilities. Compared with continual pre-training, the data scale is smaller but the risk of behavior regression is more immediate because forgetting directly affects task execution. The central design problem is therefore to balance rapid task adaptation, forgetting control, and cross-task knowledge transfer. As in traditional continual learning, existing methods can be grouped into replay-based, regularization-based, and architecture-based families, although many recent LLM methods combine multiple ingredients. Table \ref{table2} summarizes representative fine-tuning papers.

\subsection{Replay-based methods}
In the field of continual learning for LLM, replay-based methods are a key strategy designed to address the forgetting problem that models may encounter when learning new tasks. These methods assist models in retaining previously learned knowledge by preserving and utilizing data samples from old tasks, thereby facilitating continual learning. Common strategies include experience replay and generative replay. The former retains a portion of old task data samples during the training of new tasks and mixes them with new data in the training process. This approach enables models to review and reinforce their memory of old knowledge while acquiring new information. 
A representative replay-augmented continual fine-tuning objective can be abstracted as follows \cite{scialom2022fine,mok2023large}:
\begin{equation}
\mathcal{L}_{\text{replay}}=\mathbb{E}_{(x,y)\sim\mathcal{D}_t}\!\left[\mathcal{L}(f_\theta(x),y)\right]+\lambda\mathbb{E}_{(x,y)\sim\mathcal{M}_{<t}}\!\left[\mathcal{L}(f_\theta(x),y)\right],
\label{eq:replay-objective}
\end{equation}
where \(\mathcal{D}_t\) denotes the current-task data, \(\mathcal{M}_{<t}\) is replay memory or synthesized historical data, and \(\lambda\) controls the retention strength assigned to previous tasks.
All replay-based fine-tuning methods discussed below can be viewed as instances of~\eqref{eq:replay-objective}, differing primarily in how they construct the memory $\mathcal{M}_{<t}$ and weight the retention term $\lambda$. \citet{scialom2022fine} conducted the first exploration of continual learning in instruction fine-tuning, proposing a simple yet effective uniform replay with a fixed prior-data proportion. This approach helps retain earlier-task performance and enhances generalization to new instructions. DYNAINST \cite{mok2023large} extends this baseline with an entropy-based dynamic instance selection mechanism and an adaptive task replay strategy that effectively re-weights $\mathcal{M}_{<t}$ in~\eqref{eq:replay-objective} based on instance difficulty.
\cite{yin2022contintin} proposes an instruction-based continual learning method for LLMs. It first pre-trains on self-generated negative samples, then fine-tunes on normal data. During CL, it replays past task instructions instead of raw data to prevent forgetting. 
Instruction-based Continual Learning (InsCL) \cite{wang2024inscl} is a novel replay-based framework for continual instruction tuning. Leveraging task similarity computed via Wasserstein Distance over instructions, InsCL dynamically replays relevant historical data. It further proposes an Instruction Information Metric (InsInfo) to quantify instruction complexity and diversity, prioritizing high-quality samples during replay. 
 
However, sometimes the original training data may no longer be available due to privacy concerns or access restrictions. In such situations, some pseudo-sample generation techniques are proposed to create the pseudo-samples that closely resemble the distribution of original data.
SAPT \cite{zhao2024sapt} incorporates an additional pseudo-sample generator to produce synthetic samples, which helps address catastrophic forgetting. These samples assist the attention reflection module in revisiting prior knowledge, ensuring the retention of previously learned information. 
SSR \cite{huang2024mitigating} introduces Self-Synthesized Rehearsal for mitigating catastrophic forgetting in LLM continual learning. It leverages the model’s in-context learning capability to generate and iteratively refine synthetic rehearsal data, thereby eliminating the need to store real training samples while reporting strong performance and improved data efficiency relative to traditional rehearsal-based methods.

\subsection{Regularization-based methods}
Regularization-based continual learning methods represent an effective strategy for mitigating catastrophic forgetting by imposing constraints on model parameter updates. The core idea is to evaluate the importance of individual model parameters to previously learned tasks using regularization terms during the training of new tasks. Based on these evaluations, targeted constraints are applied to parameter updates. Specifically, these methods prioritize preserving parameters that are critical for earlier tasks, thereby minimizing significant changes to these parameters during gradient updates. This reduces interference with the retention of prior knowledge during new task training.
Representative regularization-based methods can be interpreted as optimizing a task loss plus an importance-weighted stability penalty \cite{mok2023large,Ke2023a}:
\begin{equation}
\mathcal{L}_{\text{reg}}=\mathcal{L}_{\text{task}}(\theta;\mathcal{D}_t)+\lambda\sum_i \Omega_i\left(\theta_i-\theta_i^{<t}\right)^2,
\label{eq:reg-objective}
\end{equation}
where \(\Omega_i\) measures the importance of parameter \(\theta_i\) to previous tasks and \(\theta_i^{<t}\) denotes its reference value before adapting to task \(t\).

A representative instantiation is the regularization branch of DYNAINST \cite{mok2023large}, which contributes the importance-weighted penalty in~\eqref{eq:reg-objective} via local-minima-inducing parameter regularization, while its replay branch (discussed above) supplies the data-side term in~\eqref{eq:replay-objective}.

\textit{Discussion.} A striking observation in the LLM continual fine-tuning literature is the scarcity of pure regularization methods compared with their prevalence in classical CL. We attribute this asymmetry to three factors. First, classical importance estimators (Fisher information in EWC \cite{kirkpatrick2017overcomingcatastrophicforgettingneural}, path integrals in SI \cite{zenke2017continuallearningsynapticintelligence}, gradient sensitivities in MAS \cite{aljundi2018memoryawaresynapticslearning}) require per-parameter statistics whose cost scales linearly with model size, becoming prohibitive at the billion-parameter regime. Second, modern LLM fine-tuning is increasingly conducted via PEFT modules (LoRA, prefix-tuning), which shrink the trainable footprint to the point where architectural isolation already provides much of the protection regularization is meant to give. Third, regularization in $\theta$-space presupposes that "important" weights are stable across tasks—an assumption weakened in over-parameterized LLMs whose loss landscape contains many degenerate solutions. We therefore view regularization not as obsolete but as best deployed \emph{inside} architectural modules (e.g., orthogonality losses on LoRA adapters, see Section~\ref{subsec:cft-arch}) rather than over the full parameter set.

\subsection{Architecture-based methods}
\label{subsec:cft-arch}
Architecture-based continual fine-tuning dynamically adjusts and expands the network structure to accommodate new tasks while preserving knowledge from prior ones. By introducing new modules, layers, or memory units, these approaches realize the \textit{isolation} mode in our knowledge-dynamics framework: previously learned weights are frozen and new tasks are absorbed by dedicated parameters, eliminating the parameter-level interference that~\eqref{eq:reg-objective} attempts to penalize \emph{a posteriori}.

A common strategy involves the use of adapters, lightweight network structures typically inserted into existing neural network layers as small, trainable modules. When learning a new task, these adapter modules are trained while the parameters of the original model remain frozen. This approach significantly reduces the number of parameters requiring adjustment, thereby lowering computational and storage costs. Additionally, the modular design of adapters enables both task-specific isolation and cross-task knowledge sharing. By incorporating adapters, models can flexibly transfer knowledge across multiple tasks while maintaining robust memory of previously learned tasks.

Multi-LoRA \cite{JIN2025112750} employs multiple LoRA modules for both target and auxiliary tasks, thereby improving the performance of continual information extraction. O-LoRA \cite{wang2023orthogonal} extends LoRA (Low-Rank Adaptation) \cite{lora} to continual learning by introducing a new LoRA module for each task in a temporal sequence, enabling efficient task-specific adaptation. To mitigate catastrophic forgetting, O-LoRA leverages the concept of orthogonal subspaces: each LoRA module is treated as an independent subspace, and an orthogonality loss is imposed across modules. This design minimizes interference from new tasks on previously acquired knowledge. By enforcing orthogonality, O-LoRA effectively isolates task representations and alleviates forgetting.

In contrast, SAPT \cite{zhao2024sapt} contends that orthogonality alone is insufficient to support knowledge transfer across tasks. To address this limitation, SAPT introduces a shared-attention framework that facilitates knowledge fusion. By leveraging attention mechanisms, SAPT enables the exchange of critical information between tasks, thereby enhancing transfer while avoiding the strict knowledge isolation inherent in orthogonality-based methods. As a result, SAPT achieves improved multi-task performance through balanced knowledge sharing and retention.

Another approach is Parameter-Efficient Fine-Tuning (PEFT), which adapts to new tasks by updating only a small subset of trainable parameters. PEFT methods offer the advantage of significantly reducing computational and storage resource requirements while maintaining efficient learning for new tasks. This makes PEFT particularly suitable for resource-constrained scenarios, where it substantially improves training efficiency.
Based on Prompt-tuning \cite{prompttuning}, \citet{razdaibiedina2023} proposes a method called progressive prompts, which freezes the model's parameters and learns a soft prompt for each task, concatenating it with prompts from previous tasks. Additionally, it introduces a more challenging continual learning task with long sequences. The method achieves strong performance in these settings.
Similarly, based on prefix-tuning \cite{li2021prefix}, \cite{wang2023rehearsal} explores a continual learning method that eliminates the need for storing historical data and relies on efficient parameter isolation. This approach performs competitively among methods that do not depend on replay. The method employs dynamic prefix tuning blocks for training and utilizes Mahalanobis distance \cite{ren2021simple} and Gaussian distributions to identify task types and select the corresponding block for new tasks.

MixLoRA \cite{li2024mixloraenhancinglargelanguage} proposes a parameter-efficient mixture-of-experts method for fine-tuning large language models. It constructs multiple LoRA-based experts within the feed-forward network block of a frozen pre-trained dense model and uses a top-k router. It also employs independent attention-layer LoRA adapters and an auxiliary load balance loss to enhance performance and address router imbalance.
\cite{du2024unlockingcontinuallearningabilities} introduces MIGU, a method for continual learning that does not require rehearsal or task labels. It updates model parameters in linear layers with large magnitudes by leveraging the observation that different tasks have unique magnitude distributions. MIGU demonstrates strong performance on a range of continual learning benchmarks and can be integrated with current CL methods to improve overall performance.
\citet{diera2024continuallearningencoderonlylanguage} proposes a DKVB architecture, which allows for localized updates by freezing the model parameters and only updating the discrete key-value pairs. It demonstrates competitive performance with lower computational costs compared to other continual learning methods. The authors also propose a generic discrete key initialization technique that is task-independent, further enhancing the model's adaptability and efficiency in various NLP tasks.
\cite{tack2024onlineadaptationlanguagemodels} proposes MAC (Memory of Amortized Contexts), an efficient online adaptation framework for LLMs. It freezes the base model and uses an amortization network to compress new data into compact modulations stored in memory. During inference, it retrieves and combines relevant modulations for adaptation—enabling knowledge retention without gradient updates. MAC outperforms fine-tuning methods in online adaptation tasks.
\cite{momeni2024continuallearningusingkernelbased} introduces KLDA for class-incremental learning (CIL), using a frozen foundation model’s features enhanced by RBF kernels and Random Fourier Features. It incrementally updates class means and a shared covariance matrix, achieving competitive performance without replay data while avoiding catastrophic forgetting and approaching the joint-training upper bound.
\cite{qiu2024continuallearningusinglarge} introduces CLOB, a verbal-prompt-based continual learning method for black-box LLMs. It introduces CIS, which incrementally updates class-wise summaries to address context limitations, achieving near-zero forgetting and strong incremental performance.
\cite{wu2024switchcitswitchingcontinualinstruction} explores SwitchCIT, which alleviates catastrophic forgetting by using a switch network to route computations to parameter-efficient tuned models. The switch network identifies tasks from instructions and routes them to specialized models fine-tuned via methods like LoRA.
\cite{araujo2024learningroutedynamicadapter} introduces L2R, which isolates the training of new PEFT modules to prevent interference with previously learned modules and employs a memory-based router to dynamically compose these modules during inference. This approach enhances generalization and performance by leveraging a small memory of previous examples to learn effective routing functions.

\subsection{Discussion: take-aways for continual fine-tuning}
\label{subsec:cft-discussion}
Three patterns emerge from the continual fine-tuning literature. \textit{First}, PEFT-based isolation via LoRA variants has become the dominant design pattern: because fine-tuning starts from a strong pre-trained base, the primary risk is behavioral regression, making isolation a cheaper defensive strategy than full-parameter regularization or replay. \textit{Second}, the small footprint of PEFT modules creates a new bottleneck—routing. Methods such as MixLoRA \cite{li2024mixloraenhancinglargelanguage}, SwitchCIT \cite{wu2024switchcitswitchingcontinualinstruction}, and L2R \cite{araujo2024learningroutedynamicadapter} invest heavily in reliable task inference at inference time; the quality of routing directly determines whether isolation remains protective or degrades into siloed modules that fail to share knowledge. \textit{Third}, the field lacks a unified benchmark covering all three method families under the same protocol, making cross-paper comparison unreliable.

\subsection{Illustrative Comparison under a Unified Protocol}
\label{subsec:cft-unified}
The methods reviewed above were proposed in different papers with different backbones, benchmarks, task orderings, and training budgets, making direct cross-paper comparison misleading. To complement the narrative survey with a concrete cross-method view, we report a supplementary comparison conducted under a unified protocol. Seven representative methods—spanning sequential baselines, regularization, and architecture/PEFT families—are evaluated with the same backbone (T5-Large), the same two benchmarks (Long Sequence Benchmark, LSB, and SuperNatural Instructions, SNI), and two task orders per benchmark, yielding four experimental conditions. The goal is not to produce a definitive ranking but to surface relative trends that are otherwise obscured by heterogeneous paper-specific setups.

\begin{figure}[!t]
\centering
\includegraphics[width=0.48\textwidth]{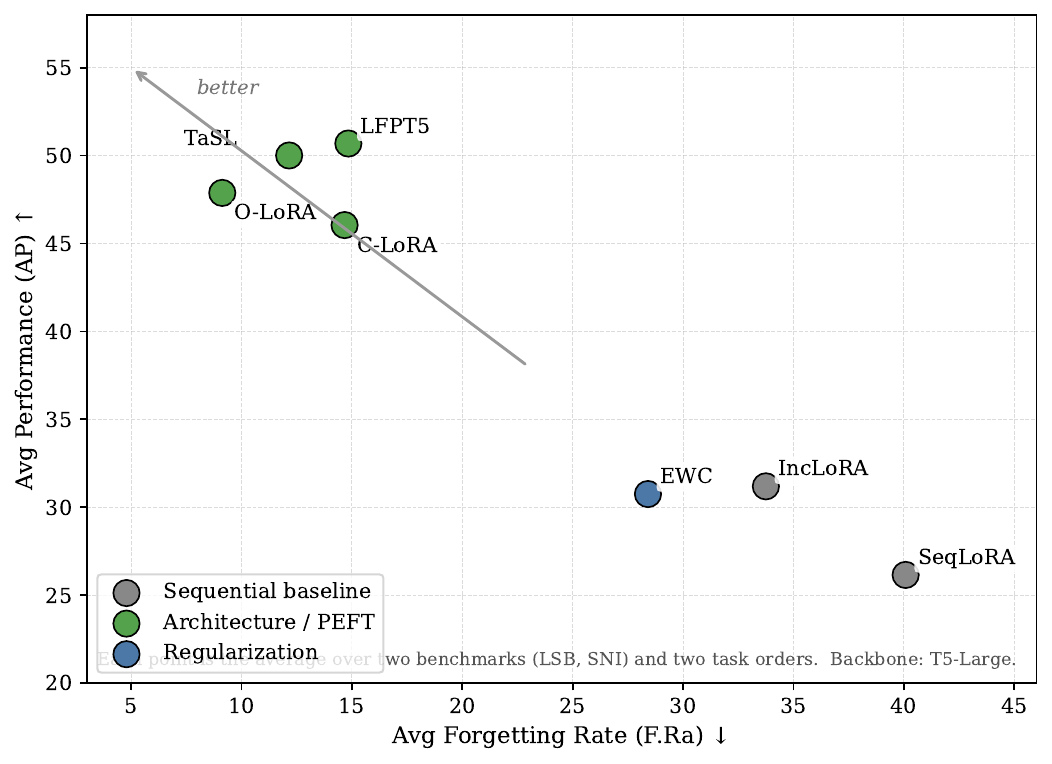}
\caption{Average performance (AP, higher is better) vs.\ average forgetting rate (F.Ra, lower is better) for seven continual fine-tuning methods under a unified T5-Large protocol (averaged over Long Sequence and SuperNI benchmarks, two task orders each). Points in the upper-left are preferable. Color indicates method family: gray = sequential baseline, blue = regularization, green = architecture/PEFT.}
\label{fig:cft-scatter}
\end{figure}

\begin{table*}[!t]
\centering
\caption{Overall results of T5-Large on two continual fine-tuning benchmarks under four task orders. LSB = Long Sequence Benchmark; SNI = SuperNatural Instructions. AP↑ = average performance (higher is better); F.Ra↓ = forgetting rate (lower is better). Best value in each column is \textbf{bold}.}
\label{tab:cft-unified}
\footnotesize
\renewcommand{\arraystretch}{1.0}
\setlength{\tabcolsep}{4.5pt}
\begin{tabular}{l|cc|cc|cc|cc|cc}
\hline
\multirow{3}{*}{Method}
  & \multicolumn{4}{c|}{Long Sequence Benchmark}
  & \multicolumn{4}{c|}{SuperNatural Instructions}
  & \multicolumn{2}{c}{Average} \\
\cline{2-11}
  & \multicolumn{2}{c|}{Order 1} & \multicolumn{2}{c|}{Order 2}
  & \multicolumn{2}{c|}{Order 3} & \multicolumn{2}{c|}{Order 4}
  & \multirow{2}{*}{AP↑} & \multirow{2}{*}{F.Ra↓} \\
\cline{2-9}
  & AP↑ & F.Ra↓ & AP↑ & F.Ra↓ & AP↑ & F.Ra↓ & AP↑ & F.Ra↓ & & \\
\hline
SeqLoRA  & 34.48 & 44.92 & 55.79 & 19.84 & 7.30 & 47.60 & 7.03 & 47.97 & 26.15 & 40.08 \\
IncLoRA  & 37.79 & 40.91 & 58.00 & 15.60 & 12.33 & 41.93 & 16.65 & 36.56 & 31.19 & 33.75 \\
LFPT5    & 66.62 & 14.57 & 67.40 & 13.20 & \textbf{39.03} & \textbf{10.87} & 29.70 & 20.72 & \textbf{50.69} & 14.84 \\
EWC      & 43.24 & 23.66 & 46.25 & 32.90 & 15.32 & 26.78 & 18.19 & 30.28 & 30.75 & 28.41 \\
O-LoRA   & 65.30 & \textbf{3.42} & 67.01 & \textbf{1.95} & 26.37 & 19.15 & 32.83 & 11.99 & 47.88 & \textbf{9.13} \\
C-LoRA   & 66.83 & 8.64 & 61.86 & 14.18 & 22.69 & 24.25 & 32.81 & 11.60 & 46.05 & 14.67 \\
TaSL     & \textbf{71.37} & 6.20 & \textbf{73.11} & 6.52 & 27.51 & 18.53 & \textbf{28.05} & 17.39 & 50.01 & 12.16 \\
\hline
\end{tabular}
\end{table*}

\textit{AP–F.Ra trade-off.} Figure~\ref{fig:cft-scatter} and Table~\ref{tab:cft-unified} jointly show that the four architecture/PEFT methods (LFPT5, O-LoRA, C-LoRA, TaSL) simultaneously dominate both AP and F.Ra over the sequential baselines (SeqLoRA, IncLoRA) and regularization (EWC), confirming that isolation-oriented designs offer a favorable stability–plasticity frontier relative to naive sequential adaptation. Within the PEFT family, a finer trade-off remains: O-LoRA achieves the lowest average F.Ra (9.13) by enforcing strict orthogonal subspace isolation, while LFPT5 achieves the highest average AP (50.69) through prompt-based task conditioning. This pattern is consistent with the claim in Section~\ref{subsec:cft-discussion} that the quality of routing and knowledge sharing—not isolation alone—determines whether a method recovers plasticity.

\textit{Order sensitivity.} Across the two Long Sequence task orders, SeqLoRA exhibits a 21-point AP swing (34.48 vs.\ 55.79), illustrating that task ordering has a disproportionate effect on simple sequential adaptation. Architecture/PEFT methods are substantially more robust: O-LoRA's AP varies by only 1.7 points across the same two orders. However, no method is fully order-insensitive: LFPT5 shows a 9.3-point gap between Orders 3 and 4 on SuperNI (39.03 vs.\ 29.70), indicating that order sensitivity is an open problem even for parameter-efficient methods when the task stream is more diverse.

\textit{No single dominant method.} Across all four conditions, no method achieves the best AP and F.Ra simultaneously. LFPT5 leads on SuperNI Order 3 and on average AP (50.69); O-LoRA leads on average F.Ra (9.13) across all orders; TaSL achieves the best per-order AP on Long Sequence. This absence of a universal winner reinforces the survey's broader point: performance in continual fine-tuning is highly benchmark- and metric-dependent, and rankings from individual papers should not be generalized beyond their specific evaluation protocols.

\textit{Scope and limitations.} This comparison is illustrative rather than exhaustive. Results are limited to T5-Large; behavior on larger decoder-only models (e.g., LLaMA-scale) may differ, particularly given the different impact of PEFT modules at higher parameter counts. The seven methods represent a cross-section of families but do not cover all approaches surveyed above. The intent is to demonstrate cross-method trends under a controlled setting, not to replace per-paper evaluations conducted under optimized conditions.

\section{Continual alignment}
Alignment \cite{alignmentsurvey} is no longer a one-off post-training step once LLM deployment spans evolving policies, domains, and user communities. Traditional RLHF pipelines assume a single alignment pass on a static preference dataset, with no mechanism to incorporate subsequent updates without re-running the full alignment pipeline—a process that becomes impractical when regulations, cultural norms, or newly discovered failure modes require targeted corrections. Continual alignment thus differs from ordinary continual fine-tuning: the alignment target itself drifts because the notion of desirable behavior is dynamic and sometimes contested.

A useful abstraction is to treat continual alignment as optimizing current preference satisfaction while preserving historically validated alignment behavior \cite{Zhang2023a,Zhang2024,li2025lifealignlifelongalignmentlarge}:
\begin{equation}
\mathcal{L}_{\text{align}}=\mathcal{L}_{\text{pref}}^{(t)}(\theta)+\lambda \mathcal{R}\!\left(\pi_\theta,\pi_{<t}\right)-\gamma \mathcal{M}_{<t},
\label{eq:align-objective}
\end{equation}
where \(\mathcal{L}_{\text{pref}}^{(t)}\) denotes the current preference objective, \(\mathcal{R}\) is a retention term anchoring the model to previously aligned policies or preference distributions, and \(\mathcal{M}_{<t}\) denotes optional memory consolidation or retrieval over historical preference traces.

\subsection{RL-free methods}
RL-free continual alignment methods instantiate~\eqref{eq:align-objective} without invoking an explicit reward model, treating the retention term $\mathcal{R}(\pi_\theta,\pi_{<t})$ as a regularizer over preference distributions rather than over scalar returns. Building on DPO~\eqref{eq:dpo}~\cite{dpo}, COPR \cite{Zhang2024} regularizes current preference learning with optimal policy targets derived from earlier stages, directly suppressing collision between newly preferred behaviors and historically aligned ones, and accompanies the first dedicated TIL and DIL benchmarks for continual preference adaptation. Beyond COPR, modular editing approaches such as LEMoE \cite{wang2024lemoeadvancedmixtureexperts} and BaFT \cite{liu2025unlockingefficientscalablecontinual} emphasize routing consistency, expert freezing, and input-conditioned composition, demonstrating that continual alignment can equally be framed as selective \textit{isolation}—editing a sparse subset of preference-relevant components rather than retraining the full policy.

\subsection{Reinforcement learning methods}
RL-based continual alignment methods retain the full RLHF pipeline but extend the PPO update in~\eqref{eq:ppo} with retention-aware terms that mirror the structure of~\eqref{eq:align-objective}. \citet{Zhang2023a} introduce continual proximal policy optimization (CPPO), which augments PPO with retention-aware reweighting to balance ongoing policy improvement and behavioral stability. LifeAlign \cite{li2025lifealignlifelongalignmentlarge} further adds memory-augmented focalized preference optimization, consolidating short-term preference traces into longer-term alignment memory—a concrete realization of the $\mathcal{M}_{<t}$ term in~\eqref{eq:align-objective}. LANCET \cite{zhang2024correctinglargelanguagemodel} complements these ideas from a corrective perspective by tracing undesirable behaviors back to influential data and refining the model accordingly, effectively performing targeted \textit{collision} cleanup.

\subsection{Preference Drift and Value Conflict}
When alignment targets evolve, the key problem is not merely forgetting but preference conflict. A response that is newly preferred in one domain or cultural setting may violate older safety assumptions or conflict with previously learned norms. This makes continual alignment qualitatively different from continual pre-training, where old knowledge is usually treated as factual memory rather than normative commitment.

Three issues are central. First, \textit{temporal weighting} determines how much new preference data should override historical alignment anchors; CPPO and COPR address this via retention-aware weighting \cite{Zhang2023a,Zhang2024}. Second, \textit{safety hierarchy} asks whether high-level safety constraints should dominate local preference updates \cite{puthumanaillam2024moralimperativeneedcontinual}. Third, \textit{domain-conditioned alignment} suggests a single globally averaged preference model is inadequate when values differ across cultures; LifeAlign \cite{li2025lifealignlifelongalignmentlarge} points toward conditionally retrieving historical preference patterns. Continual alignment is therefore not only a retention problem but a continual arbitration problem over which values should be preserved, updated, or contextually activated \cite{elelimy2025rethinkingfoundationscontinualreinforcement}.

\begin{table*}[!t]
\centering
\caption{Representative continual alignment papers. ``Type'' indicates whether the method avoids (RL-free) or uses (RL) reinforcement learning. ``Mechanism'' maps each method to the knowledge-dynamics framework: C = collision mitigation, I = isolation, F = fusion/memory integration.}
\label{tab:alignment}
\footnotesize
\renewcommand{\arraystretch}{1.0}
\begin{tabularx}{\linewidth}{>{\raggedright\arraybackslash}p{0.18\linewidth}|>{\centering\arraybackslash}p{0.10\linewidth}|>{\centering\arraybackslash}p{0.12\linewidth}|>{\centering\arraybackslash}p{0.08\linewidth}|X}
\hline
Method & Type & Base Model & Mechanism & Key Idea \\ \hline
COPR \cite{Zhang2024} & RL-free & DPO-based & C & Regularizes current preference learning with optimal-policy targets from prior stages \\
LEMoE \cite{wang2024lemoeadvancedmixtureexperts} & RL-free & MoE-based & I & Expert routing with consistency constraints to isolate preference-specific modules \\
BaFT \cite{liu2025unlockingefficientscalablecontinual} & RL-free & LLM & I & Input-conditioned module composition with expert freezing for selective alignment editing \\
CPPO \cite{Zhang2023a} & RL (PPO) & PPO-based & C & Extends PPO with retention-aware weighting to balance policy improvement and stability \\
LifeAlign \cite{li2025lifealignlifelongalignmentlarge} & RL (PPO) & LLM & F & Memory-augmented focalized preference optimization; consolidates preference traces over time \\
LANCET \cite{zhang2024correctinglargelanguagemodel} & RL-free & LLM & C & Traces undesirable behaviors to influential training data and applies targeted corrections \\ \hline
\multicolumn{5}{p{0.95\linewidth}}{\footnotesize \textit{Mechanism abbreviations}: C = collision mitigation (constraining parameter drift to preserve prior alignment); I = isolation (modular updates shielding prior policy); F = fusion/memory integration (consolidating historical preference traces).} \\
\end{tabularx}
\end{table*}

\subsection{Discussion: take-aways for continual alignment}
\label{subsec:align-discussion}
Continual alignment is the most under-studied of the three stages, and this asymmetry likely does not reflect its actual difficulty or importance. Three observations stand out. \textit{First}, the existing literature conflates two structurally distinct problems—alignment \textit{forgetting} (a previously compliant behavior degrades) and alignment \textit{conflict} (two simultaneously active norms are incompatible)—but most current methods address only the former. Preference-conflict arbitration therefore remains a largely open research direction. \textit{Second}, the handful of available methods all operate at the level of the full-model policy and rely on scalar preference scores; none yet addresses the fine-grained case where safety norms and helpfulness preferences must be kept in separate, identifiable sub-policies. Modular alignment—where safety constraints are encoded in frozen layers and helpfulness preferences are updated in lightweight adapters—appears to be a promising architectural direction for avoiding value collision without compromising adaptability. \textit{Third}, continual alignment inherits the measurement problem of conventional alignment research: reward hacking, rater drift, and Goodhart's law all apply, but in a continual setting where evaluation data from earlier stages may be unavailable or unrepresentative of current deployment contexts, these issues are compounded. Investing in dedicated longitudinal evaluation protocols is, in our view, an important infrastructure need for this sub-field.

\section{Evaluations}
The main goals of continual learning in large language models are to prevent catastrophic forgetting and facilitate knowledge transfer. Following standard continual learning evaluation protocols \cite{chen2018lifelong,ke2021achieving}, there are four main metrics in continual learning for LLMs: (1) average performance (AP), (2) forgetting rate (F.Ra), (3) forward transfer rate (FWT), and (4) backward transfer rate (BWT). Here, $a_{i,j}$ denotes the performance on task $j$ after training on task $i$.

\subsection{Average Performance}
As the main metric for evaluating the overall effectiveness of continual learning methods, the average performance, defined in~\eqref{eq:ap}, is the mean accuracy of the final model across all tasks. A high AP, however, may mask uneven task-wise behavior, which motivates the complementary metrics introduced below.
\begin{equation}
\label{eq:ap}
AP=\frac{1}{T}\sum_{i=1}^{T}a_{T,i}
\end{equation}

\subsection{Forgetting Rate}
The forgetting rate quantifies how much knowledge of old tasks has been lost after training on new tasks. As formalized in~\eqref{eq:fra}, F.Ra captures the maximum performance drop observed on each prior task between its peak and its final value. A higher F.Ra signals that learning new tasks erodes consolidated knowledge—precisely the failure mode that an effective continual learning algorithm should minimize.
\begin{equation}
\label{eq:fra}
F.Ra=\frac{1}{T-1}\sum_{i=1}^{T-1}\max_{k\in\left[i,\,T-1\right]}\left(a_{k,i}-a_{T,i}\right)
\end{equation}

\subsection{Forward Transfer Rate}
Forward transfer assesses the facilitative effect of previously acquired knowledge on the learning of new tasks. Following~\eqref{eq:fwt}, where $a_{i-1,i}$ denotes the performance on task $i$ before training on it and $b_i$ denotes the performance of a model trained solely on the $i$-th dataset, a positive FWT indicates that the continually learned model leverages prior knowledge to accelerate adaptation to incoming tasks. This is a particularly desirable property for LLMs, whose pre-training already endows them with broad priors that downstream tasks should ideally exploit rather than discard.
\begin{equation}
\label{eq:fwt}
FWT=\frac{1}{T-1}\sum_{i=2}^{T}\left(a_{i-1,i}-b_i\right)
\end{equation}

\subsection{Backward Transfer Rate}
Backward transfer measures the impact of learning new tasks on previously acquired ones, as defined in~\eqref{eq:bwt}. A positive BWT indicates that new knowledge retroactively benefits old tasks—a hallmark of genuine knowledge fusion rather than mere isolation. We argue that BWT, although less frequently reported than F.Ra, is the more discriminative indicator of whether a method achieves true cross-task synergy: methods that score well on AP but near-zero on BWT typically rely on isolation rather than integration.
\begin{equation}
\label{eq:bwt}
BWT=\frac{1}{T-1}\sum_{i=1}^{T-1}\left(a_{T,i}-a_{i,i}\right)
\end{equation}

\section{Benchmarks}
\label{sec:benchmark}
In the field of continual learning in LLMs, benchmarks play a crucial role in evaluating model performance, generalization capabilities, and resistance to catastrophic forgetting. As LLMs are increasingly applied to complex tasks and diverse domains, well-designed benchmarks are essential for assessing their ability to learn and retain knowledge in dynamic environments. Existing benchmarks typically encompass a variety of task types (e.g., classification, generation, reasoning) and simulate real-world task streams to evaluate model performance in incremental learning scenarios. Additionally, benchmarks must consider inter-task relationships, task difficulty distributions, and the diversity of evaluation metrics to reflect the strengths and limitations of models in continual learning more comprehensively. This section reviews and compares mainstream continual learning benchmarks, analyzing their design characteristics, applicable scenarios, and their contributions and limitations in evaluating LLMs.

In continual pretraining, models are typically trained on specialized vertical domains to enhance their performance in these specific areas through knowledge injection. Evaluation metrics often involve domain-specific benchmarks. Additionally, models are evaluated on general-purpose benchmarks to assess whether catastrophic forgetting has occurred in the original domain. Commonly used general-purpose benchmarks include MMLU and GSM8K, among others.

In the continual fine-tuning stage, a series of benchmarks has been proposed, typically consisting of a collection of end-task evaluations. TRACE\cite{wang2023trace} consists of several distinct challenging tasks, including domain-specific tasks, multilingual capabilities, code generation, and mathematical reasoning.

\citet{zhang2023citb} introduced CITB, a benchmark for Continual Instruction Tuning, addressing the challenge of enabling language models to continuously learn new tasks without forgetting prior knowledge. The authors curated two long task streams, InstrDialog and InstrDialog++, to systematically evaluate various continual learning methods. Their findings highlight the importance of rich natural language instructions and suggest that current continual learning methods fail to fully leverage these instructions.
\citet{jang2022towards} proposed invariantlama, updatedlama, and newlama benchmarks to evaluate knowledge retention, updating, and acquisition, along with the FUAR metric quantifying learning-forgetting trade-offs. 
\citet{razdaibiedina2023} introduce a long sequence benchmark that consists of 15 text classification tasks, which has been widely adopted.
\citet{wu2024streambench} introduces StreamBench, a benchmark for assessing LLM agents' continuous improvement via input-feedback sequences. It introduces cost-effective, multi-agent baselines that enhance performance through shared memory, advancing adaptive AI in streaming scenarios.

Table~\ref{tab:benchmarks} compares the key design properties of representative CL benchmarks across all three stages. A notable gap is the absence of a dedicated continual alignment benchmark with normative-conflict scenarios; the existing benchmarks for alignment evaluation (COPR's TIL/DIL suites \cite{Zhang2024}) are small-scale and not yet widely adopted.

\begin{table*}[!t]
\centering
\caption{Comparison of representative continual learning benchmarks. ``Stage'' indicates the primary training stage targeted; ``\# Tasks'' is the number of tasks or task streams; ``Forgetting metric'' lists the evaluation metrics provided; ``Open-ended'' indicates whether free-form generation (rather than classification) is the primary evaluation format.}
\label{tab:benchmarks}
\footnotesize
\renewcommand{\arraystretch}{1.0}
\begin{tabularx}{0.98\textwidth}{@{}lllllXl@{}}
\toprule
Benchmark & Stage & \# Tasks & Domain & Forgetting metric & Notes & Open-ended \\ \midrule
MMLU + GSM8K & CPT & 2+ & General/Math & AP, F.Ra & Widely used as CPT retention gauges & No \\
TRACE \cite{wang2023trace} & CFT & 8 & Multi-domain & AP, F.Ra & Covers code, math, multilingual, domain tasks & Yes \\
CITB / InstrDialog++ \cite{zhang2023citb} & CFT & 16 & Instructions & AP, F.Ra, BWT & Rich natural-language instruction stream & Yes \\
FUAR \cite{jang2022towards} & CFT & 3 & Knowledge & FUAR metric & Evaluates retention, update, and acquisition & No \\
Prog. Prompts (15-task) \cite{razdaibiedina2023} & CFT & 15 & Text classif. & AP, F.Ra, BWT, FWT & Long sequence; widely adopted for comparison & No \\
StreamBench \cite{wu2024streambench} & CFT/Agent & Stream & Multi-task & Improvement rate & Focuses on in-context online improvement & Yes \\
COPR TIL/DIL \cite{Zhang2024} & Alignment & 6 & Preference & AP, F.Ra & Only dedicated alignment CL benchmark & Yes \\ \bottomrule
\end{tabularx}
\end{table*}


\section{Cross-stage Synthesis}
\label{sec:synthesis}

The three training stages reviewed in this paper—continual pre-training, continual fine-tuning, and continual alignment—share a common objective (preserve while adapting) but differ substantially in their threat model, data regime, and practical constraints. Table~\ref{tab:stage-tradeoff} captures their resource trade-offs; here we synthesize the methodological divergences and convergences that cut across stage boundaries.

\begin{table*}[!t]
\centering
\caption{Three-stage $\times$ knowledge-dynamics mapping. Each cell lists representative method(s) and their primary role. C = collision mitigation; I = isolation; F = fusion. ``—'' indicates that the combination is not currently well-explored.}
\label{tab:stage-dynamics}
\footnotesize
\renewcommand{\arraystretch}{1.0}
\begin{tabularx}{0.98\textwidth}{@{}lYYY@{}}
\toprule
 & \textbf{Collision mitigation} & \textbf{Isolation} & \textbf{Fusion} \\
\midrule
\textbf{Cont. pre-training}
  & Data augmentation~\cite{Guo2024,Xie2024}; process optimization~\cite{Jiang2024a} (C)
  & Adapter plugins~\cite{Ke2022}; DGA/DAS~\cite{Ke2023a}; LLaMA-MoE~\cite{Zhu2024} (I)
  & Rehearsal data mixing~\cite{fujii2024continual,Cui2024,Que2024} (F) \\
\midrule
\textbf{Cont. fine-tuning}
  & Regularization inside LoRA adapters~\cite{wang2023orthogonal} (C)
  & O-LoRA~\cite{wang2023orthogonal}; Progressive Prompts~\cite{razdaibiedina2023}; L2R~\cite{araujo2024learningroutedynamicadapter} (I)
  & SAPT shared attention~\cite{zhao2024sapt}; InsCL replay~\cite{wang2024inscl}; SSR~\cite{huang2024mitigating} (F) \\
\midrule
\textbf{Cont. alignment}
  & COPR~\cite{Zhang2024}; CPPO~\cite{Zhang2023a}; LANCET~\cite{zhang2024correctinglargelanguagemodel} (C)
  & LEMoE~\cite{wang2024lemoeadvancedmixtureexperts}; BaFT~\cite{liu2025unlockingefficientscalablecontinual} (I)
  & LifeAlign memory consolidation~\cite{li2025lifealignlifelongalignmentlarge} (F); — \\
\bottomrule
\end{tabularx}
\end{table*}

\subsection{Divergences across stages}
\textit{Data scale and supervision.} CPT operates on unlabeled corpora at the trillion-token scale, where per-sample gradient signals are diffuse and forgetting manifests slowly but persistently. CFT works with curated instruction datasets at the million-sample scale, where task identity is explicit and forgetting is immediate and task-specific. Continual alignment trains on pairwise preference feedback at the thousand-example scale, where the supervision signal encodes normative rather than factual content. These three regimes call for qualitatively different detection and intervention strategies.

\textit{Nature of what is forgotten.} CPT forgets factual and linguistic representations; the forgotten content is relatively stable and objective, making evaluation straightforward. CFT forgets behavioral patterns—instruction following on particular task formats—which are more brittle but also more recoverable through replay. Continual alignment forgets normative commitments, which are neither stable nor objective, making both detection and recovery fundamentally harder.

\textit{Method efficacy by family.} Rehearsal is most commonly and successfully applied in CPT, where the only cost is storage and mixing logistics. Architecture-based isolation is most commonly adopted in CFT, where PEFT modules provide lightweight per-task protection. For continual alignment, neither family is yet dominant, and the field is still in an exploratory phase.

\subsection{Convergences and shared principles}
Despite these differences, three design principles recur across all three stages and can be regarded as universal heuristics for LLM continual learning.

\textit{Sparse updates protect more than dense updates.} Across CPT (adapter plugins \cite{Ke2022}), CFT (LoRA isolation \cite{wang2023orthogonal}), and alignment (modular editing \cite{wang2024lemoeadvancedmixtureexperts}), restricting gradient flow to a small subset of parameters consistently reduces forgetting without proportional sacrifices in plasticity.

\textit{Memory and retrieval complement parametric updates.} Whether the memory stores pre-training tokens (CPT rehearsal), historical instruction examples (CFT replay), or past preference traces (LifeAlign \cite{li2025lifealignlifelongalignmentlarge}), providing the optimizer with access to historical data at training time is the most reliable forgetting-mitigation strategy currently available.

\textit{Stage-aware initialization accelerates adaptation.} PIT \cite{Jiang2024a} pre-conditions CPT with instruction-tuning; SSR \cite{huang2024mitigating} uses the model itself to generate synthetic rehearsal data; COPR \cite{Zhang2024} initializes alignment with prior optimal policies. In each case, using the previous stage's outputs as the starting point for the next reduces the magnitude of corrective updates required and shrinks the collision footprint.

\subsection{Open coordination problem}
A key observation from this cross-stage perspective is that the three stages are currently treated as sequential and independent pipeline steps, but in practice they interact: a poorly designed CPT may amplify forgetting during subsequent CFT, and misaligned fine-tuning may corrupt alignment stability. We argue that the field would benefit from end-to-end continual learning protocols that optimize the full CPT $\to$ CFT $\to$ Alignment pipeline jointly, rather than solving each stage in isolation. One promising direction is joint objective design where CPT corpus selection, CFT task ordering, and alignment preference weighting are co-optimized against a unified retention-transfer metric, ensuring that decisions made at the pre-training stage do not inadvertently narrow the hypothesis space available to subsequent fine-tuning and alignment. This represents, in our view, an important architectural shift for moving from isolated CL methods toward a practical lifelong LLM infrastructure.

\section{Challenges and opportunities}
\label{sec:challenges}
\subsection{Challenges}
\subsubsection{Catastrophic Forgetting at Scale}
Catastrophic forgetting—sharp performance degradation on prior tasks when training on new ones—remains the central obstacle, and its severity is amplified at the LLM scale. Classical remedies such as EWC \cite{kirkpatrick2017overcomingcatastrophicforgettingneural} rely on per-parameter importance statistics whose memory and compute cost grow linearly with model size, rendering them infeasible for billion-parameter models. Replay-based approaches face the twin burdens of storage and privacy: retaining real samples is increasingly restricted by data-governance regulations, while synthetic replay is still insufficiently faithful to original distributions in open-ended generation tasks. We argue that forgetting at the LLM scale is not a scaled-up version of the classical problem but a qualitatively different one, because the over-parameterized loss landscape offers many degenerate paths that can satisfy new objectives without correcting old ones—making it hard to detect forgetting before it becomes severe.

\subsubsection{Scalability and Efficiency}
Continual learning methods designed for small classifiers often impose per-task overheads (growing architectures, fresh LoRA modules, updated importance matrices) that accumulate over long task sequences. The field lacks a principled account of how these costs scale with the number of tasks, the size of the base model, and the diversity of the task stream. Without such an account, claims of "efficient" CL are difficult to compare across papers.

\subsubsection{Limited Knowledge Transfer}
Most existing methods prioritize stability (preventing forgetting) over plasticity (facilitating transfer). Forward transfer—where prior knowledge actively accelerates learning of new tasks—is rarely reported and even more rarely optimized. We view this imbalance as a missed opportunity: LLMs already possess rich prior representations that should, in principle, enable steep learning curves on related tasks. Closing the gap between theoretical potential and observed transfer rates is, in our view, one of the highest-value research directions in this field.

\subsubsection{Evaluation and Benchmark Gaps}
As detailed in Section~\ref{sec:benchmark}, existing benchmarks cover continual pre-training and fine-tuning reasonably well but leave continual alignment nearly unaddressed. More critically, most benchmarks evaluate a fixed short task sequence (typically 5–15 tasks), making it difficult to assess long-horizon behavior. Pseudo-sample generation has emerged as a promising approach to mitigate data-privacy concerns in replay \cite{lee2024llm2llm}; techniques from federated learning \cite{li2024synergizingfoundationmodelsfederated} offer complementary tools for privacy-preserving replay, but their integration with continual learning evaluation remains nascent.

\subsubsection{Cross-Modal and Multi-Stage Generalization}
As LLMs increasingly operate in multimodal settings, continual learning must address cross-modal alignment drift: the gradual desynchronization of representations across modalities when new data arrives for only one modality. Real deployments further involve sequential stage transitions (CPT $\to$ CFT $\to$ Alignment), where interventions at one stage propagate as initialization bias into subsequent stages. The field lacks systematic methods for diagnosing these cross-stage interactions, meaning gains at one stage can be silently erased by the next—a structural challenge with no adequate solution yet.

\subsection{Opportunities}
Despite recent progress, many promising research opportunities remain open, including but not limited to the following.

\subsubsection{Improving Performance}
Developing more effective methods to enhance the efficiency, scalability, and robustness of continual learning remains a central challenge. Several studies approach catastrophic forgetting through principled geometric analyses—for example, ProNC \cite{wang2025rethinkingcontinuallearningprogressive} uses ETF-structured prototypes to preserve geometric separability across tasks. A key open direction is understanding why the ``stability gap'' occurs and how the loss landscape evolves under sequential domain shifts, which could support more principled continual-update methods.

\subsubsection{Multimodal Continual Learning}
Continual learning research has predominantly focused on NLP, while multimodal continual learning (MM-CL) remains underexplored. MM-CL must address not only catastrophic forgetting within each modality but also cross-modal alignment drift, where representations become inconsistent as new data arrives for only one modality. Recent work \cite{zeng2025modalpromptefficientmultimodalcontinual,cai2025empoweringlargelanguagemodel,li2024atlasadapterbasedmultimodalcontinual,liu2025cclip} proposes efficient continual instruction-tuning frameworks for multimodal models, and a key open question is whether modality-invariant prompts can decouple modality-specific drift from task-semantic evolution.

\subsubsection{Reinforcement Learning Approaches}
Recent studies suggest that LLMs trained through reinforcement learning can retain acquired knowledge more effectively, indicating the potential of RL-based strategies for advancing continual learning. Leveraging RL in CL enables models to actively interact with the learning environment, facilitating adaptive knowledge acquisition and long-term retention. \citet{pan2025surveycontinualreinforcementlearning} provides a comprehensive survey of RL methods in the context of continual learning; \cite{hu2025continualknowledgeadaptationreinforcement,wei2025sftsecondrlupt} demonstrate that RL can enhance knowledge transfer in sequential tasks. An important direction for future work is to examine whether reward signals can serve as an implicit curriculum, especially in online CL settings where task boundaries are unclear.

\subsubsection{Semi-Parametric Methods}
Most traditional continual learning approaches rely on fully parametric models that continuously update weights to incorporate new knowledge. Semi-parametric methods have recently emerged as a promising alternative \cite{du2023staticdynamiccontinuallearning,zheng2025lifelonglearninglargelanguage}, combining parametric updates with non-parametric memory mechanisms to achieve a more effective balance between plasticity and stability. Semi-parametric frameworks often employ external memory modules, retrieval-augmented components, or agent-based architectures that selectively recall relevant past information without overwriting model parameters. These designs may be especially attractive for long-horizon continual learning, because the non-parametric component can absorb fast factual updates while the parametric component remains a relatively stable competence store.

\subsubsection{Towards a General CL Agent}
A compelling long-term direction is a general continual learning agent that combines external retrieval, selective parameter updates, and continual alignment within one loop. Retrieval-augmented memory would absorb recent or provenance-sensitive knowledge without immediately overwriting model parameters, while repeated or high-value patterns would be selectively internalized through modular or parameter updates \cite{tack2024onlineadaptationlanguagemodels,li2025cmt,zheng2025lifelonglearninglargelanguage}. Continual alignment would govern how these evolving capabilities are prioritized under changing safety requirements, mirroring lifelong human learning. A central open problem in this setting is stage coordination: when to retrieve, when to internalize, and when to re-align.

\subsubsection{Online Continual Learning}
In real-world environments, task boundaries are often ambiguous or non-existent. Designing algorithms that learn from streaming data while mitigating catastrophic forgetting is a key frontier for practical deployment. Recent studies \cite{li2025cmt,bidaki2025onlinecontinuallearningsystematic} explore OCL frameworks that update models continuously without revisiting past tasks, likely requiring retrieval-based mechanisms for time-sensitive knowledge while reserving gradient updates for validated information.


\section{Conclusion}
Moving beyond static models requires treating continual learning not as an isolated toolbox but as an evolving framework for large language models across pre-training, fine-tuning, and alignment. We have organized the field around these three stages, highlighted the knowledge dynamics that govern collision, isolation, and fusion, and argued that continual alignment introduces a distinct problem of preference drift and value conflict rather than simple task transfer.

\textit{Stage-level findings.} For continual pre-training, rehearsal is a strong baseline—mixing 5–15\% of prior-domain data with new corpora consistently preserves general capabilities, though the optimal ratio remains setting-dependent. The "stability gap"—an initial performance dip before recovery—remains theoretically unexplained. For continual fine-tuning, PEFT-based isolation has become the dominant strategy; the main open subproblem is reliable task routing at inference, which determines whether isolated modules remain accessible and composable. For continual alignment, the distinction between alignment forgetting and preference conflict has only recently been articulated, and its practical consequences remain poorly characterized.

\textit{Cross-stage perspective.} The three stages share a common design pressure: sparse, memory-augmented, stage-aware updates often outperform dense full-parameter rewrites, but they currently operate as independent pipeline steps. As argued in Section~\ref{sec:synthesis}, the field would benefit from end-to-end protocols that optimize the full CPT $\to$ CFT $\to$ Alignment pipeline jointly, rather than solving each stage in isolation. The interaction effects between stages remain underappreciated: CPT geometry shapes the landscape on which CFT operates, and CFT behavior determines the difficulty of subsequent alignment.

\textit{Priorities for future work.} Three priorities stand out. First, a principled theory of forgetting at the LLM scale—accounting for over-parameterization, the stability gap, and sequential domain-shift geometry—is needed to move beyond empirical heuristics. Second, longitudinal benchmarks covering long task sequences (50+ tasks), multiple training stages, and alignment-specific metrics would replace the fragmented evaluations currently used; Section~\ref{subsec:cft-unified} shows how even a minimal unified protocol surfaces behaviors—order sensitivity, the AP–F.Ra trade-off—invisible from per-paper rankings. Third, a general CL agent combining retrieval, selective internalization, and continual alignment offers a long-term path toward LLMs that evolve safely with new knowledge.




\bibliography{reference}

 



\end{document}